\documentclass[twocolumn,twoside]{IEEEtran}

\usepackage{amsmath,amssymb,amsthm,epsfig,dsfont,color,subfigure,empheq}
\usepackage{enumerate,stfloats,url,algpseudocode,algorithm}

\DeclareMathOperator{\rank}{rank}

\DeclareMathOperator{\trace}{Tr}
\DeclareMathOperator{\vectorize}{vec}

\DeclareMathOperator{\subjectto}{s.to}

\newtheorem{proposition}{Proposition}
\newtheorem{lemma}{Lemma}
\newtheorem{corollary}{Corollary}
\newtheorem{theorem}{Theorem}

\theoremstyle{remark}\newtheorem{remark}{Remark}

\begin{document}
\bstctlcite{IEEEexample:BSTcontrol}

\title{Electricity Market Forecasting\\
via Low-Rank Multi-Kernel Learning}

\author{Vassilis Kekatos,~\IEEEmembership{Member,~IEEE,} Yu Zhang,~\IEEEmembership{Student Member,~IEEE,} and Georgios B.
Giannakis*,~\IEEEmembership{Fellow,~IEEE} %
\thanks{Work in this paper was supported by the Inst. of Renewable Energy and the Environment (IREE) under grant no. RL-0010-13, Univ. of Minnesota, and NSF Grant ECCS-1202135. The authors are with the ECE Dept., University of Minnesota, Minneapolis, MN 55455, USA. Emails:\{kekatos,zhan1220,georgios\}@umn.edu}}


\maketitle

\begin{abstract}
The smart grid vision entails advanced information technology and data analytics to enhance the efficiency, sustainability, and economics of the power grid infrastructure. Aligned to this end, modern statistical learning tools are leveraged here for electricity market inference. Day-ahead price forecasting is cast as a low-rank kernel learning problem. Uniquely exploiting the market clearing process, congestion patterns are modeled as rank-one components in the matrix of spatio-temporally varying prices. Through a novel nuclear norm-based regularization, kernels across pricing nodes and hours can be systematically selected. Even though market-wide forecasting is beneficial from a learning perspective, it involves processing high-dimensional market data. The latter becomes possible after devising a block-coordinate descent algorithm for solving the non-convex optimization problem involved. The algorithm utilizes results from block-sparse vector recovery and is guaranteed to converge to a stationary point. Numerical tests on real data from the Midwest ISO (MISO) market corroborate the prediction accuracy, computational efficiency, and the interpretative merits of the developed approach over existing alternatives.
\end{abstract}

\begin{keywords}
Kernel-based learning; nuclear norm regularization; multi-kernel learning; graph Laplacian; commercial pricing nodes; block-coordinate descent; low-rank.
\end{keywords}

\section{Introduction}\label{sec:intro}
Forecasting electricity prices is an important decision making tool for market participants~\cite{AmHe06}. Conventional and particularly renewable asset owners plan their trading and bidding strategies according to pricing predictions. Moreover, independent system operators (ISOs) recently broadcast their own market forecasts to proactively relieve congestion~\cite{ERCOT}. At a larger geographical and time scale, electricity price analytics based solely on publicly available data rather than physical system modeling are pursued by government services to identify ``national interest transmission congestion corridors''~\cite{DOE_corridors}.

In a generic electricity market setup, an ISO collects bids submitted by generator owners and utilities~\cite{SPM2013},~\cite{KirschenStrbac}. Compliant with network and reliability constraints, the grid is dispatched in the most economical way. Following power demand patterns, electricity prices exhibit cyclo-stationary motifs over time. More importantly and due to transmission limitations, cheap electricity cannot be delivered everywhere across the grid. Out-of-merit energy sources have to be dispatched to balance the load. Hence, congestion together with heat losses lead to spatially-varying energy prices, known as locational marginal prices (LMPs) \cite{KirschenStrbac}, \cite{ExpConCanBook}.

Schemes for predicting electricity prices proposed so far include time-series analysis approaches based on auto-regressive (integrated) moving average models and their generalizations~\cite{Contreras03}, \cite{Garcia05}. \textcolor{black}{However, these models are confined to linear predictors, whereas markets involve generally nonlinear dependencies. To account for nonlinearities, artificial intelligence approaches, such as fuzzy systems and neural networks, have been investigated \cite{Zhang03}, \cite{Li07}, \cite{WuShahi10}. Hidden Markov models have been also advocated \cite{Gonzalez05}.} A nearest neighborhood method was suggested in \cite{Lora07}. Market clearance was solved as a quadratic program and forecasts were extracted based on the most probable outage combinations in \cite{ZhTeCh11}. Reviews on electricity price forecasting and the associated challenges can be found in \cite{AmHe06} and \cite{ShYaLi02}.

Different from existing approaches where predictors are trained on a per-node basis, a framework for learning the entire market is pursued in this work. Building on collaborative filtering ideas, market forecasting is cast as a learning task over all nodes and several hours~\cite{Aber09}, \cite{Argy09}. Leveraging market clearing characteristics, prices are modeled as the superposition of several rank-one components, each capturing particular spatio-temporal congestion motifs. Distinct from \cite{KeVeLiGia13}, \emph{low-rank} kernel-based learning models are developed here.

A systematic kernel selection methodology is the second contribution of this paper. Due to the postulated decomposition, different kernels must be defined over nodes and hours. Our novel analytic results extend kernel learning tools to low-rank multi-task models \cite{Mich05}, \cite{Gonen11}, \cite{AlRoLa12}. By viewing market extrapolation as learning over a graph, the commercial pricing network is surrogated here via balancing authority connections and meaningful graph Laplacian-based kernels are provided.

An efficient algorithm for solving the computationally demanding optimization involved is our third contribution. Although the problem is jointly non-convex, per block optimizations entail convex yet non-differentiable costs which are tackled via a block-coordinate descent approach. Leveraging results from (block) compressed sensing~\cite{Puig11}, the resultant algorithm boils down to univariate minimizations, exploits the Kronecker product structure, and is guaranteed to converge to a stationary point of the resultant optimization problem. Forecasting results on the MISO market over the summer of 2012 corroborate the accuracy, interpretative merit, and the computational efficiency of the novel learning model. 

\emph{Notation.} Lower- (upper-) case boldface letters denote column vectors (matrices); calligraphic letters stand for sets. Symbols $(\cdot)^{\top}$ and $\otimes$ denote transposition and the Kronecker product, respectively. The $\ell_2$-norm of a vector is denoted by $\|\mathbf{a}\|_2$, $\|\mathbf{A}\|_F$ is the Frobenius matrix norm, and $\mathbb{S}_{++}^N$ is the set of $N\times N$ positive definite matrices. The operation $\vectorize(\mathbf{A})$ turns matrix $\mathbf{A}$ to a vector by stacking its columns, and $\trace(\mathbf{A})$ denotes its trace. The property $\vectorize(\mathbf{AXB}) = \left(\mathbf{B}^{\top} \otimes \mathbf{A}\right)\vectorize(\mathbf{X})$ will be needed throughout.

The paper outline is as follows. Electricity market forecasting is formulated in Sec.~\ref{sec:work}, where the novel approach is presented. A block-coordinate descent algorithm is detailed in Sec.~\ref{sec:bcd}. Kernel design and forecasting results on the MISO market are in Sec.~\ref{sec:simulations}. The paper is concluded in Sec.~\ref{sec:conclusions}.

\section{Problem Statement and Formulation}\label{sec:work}

\subsection{Preliminaries on Kernel-Based Learning}\label{subsec:background}
Given pairs $\{(x_n,z_n)\}_{n=1}^N$ of features $x_n$ belonging to a measurable space $\mathcal{X}$ and target values $z_n\in \mathbb{R}$, kernel-based learning aims at finding a relationship $f:\mathcal{X}\rightarrow\mathbb{R}$ with $f$ belonging to the linear function space
\begin{equation}\label{eq:family}
\mathcal{H}_{\mathcal{K}}:=\left\{f(x)=\sum_{n=1}^{\infty} K(x,x_n) a_n,~a_n\in\mathbb{R}\right\}
\end{equation}
defined by a preselected kernel (basis) $K:\mathcal{X}\times \mathcal{X}\rightarrow \mathbb{R}$ and corresponding coefficients $a_n$. When $K(\cdot,\cdot)$ is a symmetric positive definite function, then $\mathcal{H}_{\mathcal{K}}$ becomes a reproducing kernel Hilbert space (RKHS) whose members have a finite norm $\|f\|_{\mathcal{K}}^2:= \sum_{n=1}^{\infty} \sum_{m=1}^{\infty} K(x_n,x_m) a_n a_m$~\cite{Aronszajn}.

Viewed either from a Bayesian estimation perspective, or as a function approximation task, learning $f$ can be posed as the regularization problem~\cite{Hastie}, \cite{BaGia13}
\begin{equation}\label{eq:fnreg}
\hat{f}:=\arg\min_{f\in\mathcal{H}_{\mathcal{K}}} ~ \sum_{n=1}^N (z_n-f(x_n))^2 + \mu\|f\|_{\mathcal{K}}.
\end{equation}
The least-squares (LS) fitting component in \eqref{eq:fnreg} captures the designer's reliance on data, whereas the regularizer $\|f\|_{\mathcal{K}}$ constraints $f\in\mathcal{H}_{\mathcal{K}}$ and facilitates generalization over unseen data. The two components are balanced through the parameter $\mu>0$, which is typically tuned via cross-validation~\cite{Hastie}.

Finding $\hat{f}$ requires solving the functional optimization in \eqref{eq:fnreg}. Fortunately, the celebrated Representer's Theorem asserts that $\hat{f}$ admits the form $\hat{f}(x)=\sum_{n=1}^N K(x,x_n)\hat{a}_n$~\cite{Hastie}. Hence, the sought $\hat{f}$ can be characterized by the coefficient vector $\hat{\mathbf{a}}:=[\hat{a}_1\cdots \hat{a}_N]^{\top}$. Upon defining the kernel matrix $\mathbf{K}\in \mathbb{S}_{++}^{N}$ having entries $[\mathbf{K}]_{n,m}:=K(x_n,x_m)$, the vector $\mathbf{z}:=[z_1 \cdots z_N]^{\top}$, and the norm $\|\mathbf{a}\|_{\mathbf{K}}^2:= \mathbf{a}^{\top}\mathbf{K}\mathbf{a}$; solving \eqref{eq:fnreg} is equivalent to the vector optimization
\begin{equation}\label{eq:vecreg}
\hat{\mathbf{a}}:=\arg\min_{\mathbf{a}} ~ \|\mathbf{z}-\mathbf{K}\mathbf{a}\|_2^2 + \mu\|\mathbf{a}\|_{\mathbf{K}}.
\end{equation}
Building on kernel-based learning, novel models pertinent to electricity market forecasting are pursued next.

\subsection{Low-Rank Learning}\label{sec:approach}
Consider a whole-sale electricity market over a set $\mathcal{N}$ of $N$ commercial pricing nodes (CPNs) indexed by $n$. In a day-ahead market, locational marginal prices (LMPs) correspond to the cost of buying or selling electricity at each CPN and over one-hour periods for the following day~\cite{Ott03},~\cite{ExpConCanBook}.

\color{black}

Viewing market forecasting as an inference problem, day-ahead LMPs are the target variables to be learned. Explanatory variables (features) can be any data available at the time of forecasting and believed to be relevant to the target variables. Due to the spatiotemporal nature of the problem, features can be either related to a CPN (nodal features), or a specific market hour (time features). Candidate nodal features could be the node type (generator, load, interface to another market); the generator technology (coal, natural gas, nuclear, or hydroelectric plant, wind farm); CPN's geographical location; and the balancing authority controlling the node. Vector $\mathbf{x}_n$ collects the features related to the $n$-th CPN.

Vector $\mathbf{y}_t$ comprises the features related to a market period $t$, say 3pm on August 1st, 2012. Candidate features could be:
\begin{itemize}
\item the 3pm LMPs from past days;
\item load estimates (issued per balancing authority, region, and/or the market footprint);
\item weather forecasts (e.g., temperature, humidity, wind speed, and solar radiation at selected locations);
\item outage capacity (capacity of generation units closed for maintenance);
\item timestamp features (hour of the day, day of the week, month of the year, holiday) to capture peak demand hours on weekdays as well as heating and cooling patterns;
\item scheduled power imports and exports to other markets.
\end{itemize}
Note that $\mathbf{y}_t$ is shared across CPNs. Weather forecasts across major cities or renewable energy sites affect several CPNs, while capacity outages, regional load estimates, and timestamps relate to the whole market. Moreover, the location of CPNs may be unknown.

A generic approach could be to predict every single-CPN price given $\mathbf{y}_t$ and the observed LMPs. Such an approach would train $N$ separate prediction models with identical feature variables. However, locational prices are not independent. They are determined over a transmission grid having capacity and reliability limitations \cite{SPM2013},~\cite{KeGiBa14}. Leveraging this network-imposed dependence, market forecasting is uniquely interpreted here as learning over a graph; see e.g., \cite{Kolaczyk}. Energy markets may change significantly due to lasting transmission and generation outages, or shifts in oil or gas markets. That is why the market is considered to be stationary only over the $T$ most recent time periods, which together with the sought next 24 hours comprise the set $\mathcal{T}$. The market could be then thought of as a function $p:\mathcal{N}\times \mathcal{T}\rightarrow \mathbb{R}$ to be inferred.

We postulate that the price at node $n$ and time $t$ denoted by $p(n,t)$ belongs to the RKHS defined by the tensor product kernel $K_{\otimes}\left((n,t),(n',t')\right):=K(n,n')G(t,t')$, where $K:\mathcal{N}\times \mathcal{N}\rightarrow \mathbb{R}$ and $G:\mathcal{T}\times \mathcal{T}\rightarrow \mathbb{R}$ are judiciously selected kernels over nodes and hours. The tensor product kernel is a valid kernel and has been used in collaborative filtering and multi-task learning \cite{Aber06}, \cite{Aber09}, \cite{Mich05}, \cite{Kolt10}. All functions in this RKHS, denoted by set $\mathcal{P}$, can be alternatively represented as~\cite{Aronszajn},~\cite{Aber09}
\begin{equation}\label{eq:F}
\mathcal{P}=\left\{ p(n,t)=\sum_{r=1}^R f_r(n) g_r(t), ~f_r\in \mathcal{H}_{K},~g_r\in \mathcal{H}_{G}\right\}
\end{equation}
where $\mathcal{H}_{\mathcal{K}}$ and $\mathcal{H}_{\mathcal{G}}$ are the RKHSs defined respectively by $K$ and $G$, while the number of summands $R$ is possibly infinite. Note that the decomposition in \eqref{eq:F} is not unique~\cite{Aronszajn}. Similar to \eqref{eq:fnreg} and upon arranging observed prices in $\mathbf{Z}\in\mathbb{R}^{N\times T}$, the market function $p(n,t)$ could be inferred via
\begin{equation}\label{eq:regularization2}
\min_{p\in\mathcal{P}}~\|\mathbf{Z}-\mathbf{P}\|_F^2 + \mu\|p\|_{\mathcal{K}_{\otimes}}
\end{equation}
where $\mathbf{P}\in\mathbb{R}^{N\times T}$ has entries $[\mathbf{P}]_{n,t}=p(n,t)$, $\|p\|_{\mathcal{K}_{\otimes}}$ is the norm in $\mathcal{P}$ [cf.~\eqref{eq:family}], and $\mu>0$ is a regularization parameter. Notice the notational convention that when $n$ and $t$ are used as arguments, the function depend on $\mathbf{x}_n$ and $\mathbf{y}_t$, respectively. In other words, $p(n,t)=p(\mathbf{x}_n,\mathbf{y}_t)$, $K(n,n')=K(\mathbf{x}_n,\mathbf{x}_{n'})$, and $G(t,t')=G(\mathbf{y}_t,\mathbf{y}_{t'})$.

\color{black}

The key presumption here is that $p(n,t)$ is practically the superposition of relatively few components $p_r(n,t):=f_r(n)g_r(t)$: At a specific $t$, usually only a few transmission lines are congested, i.e., have reached their rated power capacity~\cite{ExpConCanBook,SPM2013}.\footnote{This fact is exploited in \cite{KeGiBa14} to reveal the topology of the underlying power grid by using only publicly available real-time LMPs.} Each $p_r$ corresponds to the pricing pattern observed whenever a specific congestion scenario occurs. Yet spatial effects are modulated by time. For example, congestion typically occurs during peak demand or high-wind periods. Moreover, due to generator ramp constraints, demand periodicities, and lasting transmission outages; pricing motifs tend to iterate over time instances with similar characteristics, e.g., the same hour of the next day or week. These specifications not only justify using the tensor product kernel $K_{\otimes}$, but they also hint at a relatively small $R$ in \eqref{eq:F}.

\color{black}
To facilitate parsimonious modeling of $p(n,t)$ using a few $p_r(n,t)$ components, instead of regularizing by $\|p\|_{K_{\otimes}}$ [cf.~\eqref{eq:regularization2}], the \emph{trace norm} $\|p\|_*$ could be used:
\begin{equation}\label{eq:regularization2.5}
\min_{p\in\mathcal{P}}~\|\mathbf{Z}-\mathbf{P}\|_F^2 + \lambda\|p\|_*
\end{equation}
for some $\lambda>0$. For the definition of trace norm see \cite{Aber06}. In \cite{Aber06}, it is also shown that for every function $p\in\mathcal{P}$, its $\|p\|_*$ can be alternatively expressed as
\begin{align}\label{eq:tracenorm}
\|p\|_*=\min_{\{f_r,g_r\}} ~&~\frac{1}{2}\left(\sum_{r=1}^R\|f_r\|_{\mathcal{K}}^2+ \sum_{r=1}^R\|g_r\|_{\mathcal{G}}^2\right)\\ 
\subjectto~&~p=\sum_{r=1}^R f_r g_r, ~f_r\in \mathcal{H}_{K},~g_r\in \mathcal{H}_{G}.\nonumber
\end{align}

Regularizing by $\|p\|_*$ is known to favor low-rank models~\cite{Aber09,ReFaPa10}. Nevertheless, in this work we advocate regularizing by the square root of $\|p\|_*$ to critically enable kernel selection (cf.~Section~\ref{subsec:mkl}) and to derive efficient algorithms (cf.~Section~\ref{sec:bcd}). In detail, market inference is posed here as the regularization problem:
\begin{equation}\label{eq:regularization3}
\min_{p\in\mathcal{P}}~\|\mathbf{Z}-\mathbf{P}\|_F^2 + \mu\sqrt{\|p\|_*}
\end{equation}
for some $\mu>0$. The connection between \eqref{eq:regularization2.5} and \eqref{eq:regularization3} can be understood by the next proposition proved in Appendix~\ref{subsec:subset}.

\begin{proposition}\label{pro:subset}
Let $p_{\mu}^*$ be a function minimizing \eqref{eq:regularization3} for some $\mu>0$. There exists $\lambda_{\mu}>0$, such that $p_{\mu}^*$ is also a minimizer of \eqref{eq:regularization2.5} for $\lambda=\lambda_{\mu}$.
\end{proposition}

Albeit Proposition~\ref{pro:subset} does not provide an analytic expression for $\lambda_{\mu}$, it asserts that every minimizer of \eqref{eq:regularization3} is a minimizer for \eqref{eq:regularization2.5} too for an appropriate $\lambda$. Thus, the functions minimizing \eqref{eq:regularization3} are expected to be decomposable into a few $p_r$. Numerical tests indicate that \eqref{eq:regularization3} favors low-rank minimizers indeed. 

\color{black} 
Given that \eqref{eq:regularization3} admits low-rank minimizers anyway, its feasible set could be possibly restricted to a $\mathcal{P}$ defined by \eqref{eq:F} but for a finite and relatively small $R_0$. If the $p$ minimizing \eqref{eq:regularization3} over this restricted feasible set turns out to be of rank smaller than $R_0$, the restriction comes at no loss of optimality. Throughout the rest of the paper, \eqref{eq:regularization3} will be solved for a finite $R$. Similar approaches have been developed for low-rank matrix completion \cite{BaGia13}, collaborative filtering \cite{Aber09}, and multi-task learning \cite{Mich05}, \cite{Kolt10}.

To leverage the low-rank model in solving \eqref{eq:regularization3}, the following result, proved in Appendix~\ref{subsec:sqrt_proof}, is needed:
\begin{lemma}\label{le:sqrt}
For every $p\in\mathcal{P}$, it holds \textcolor{black}{$\sqrt{\|p\|_*}=h(p)$}, where
\begin{align}\label{eq:sqrt}
h(p):=\min_{\{f_r,g_r\}}& 
\frac{1}{2}\left[\left(\sum_{r=1}^R \|f_r\|_{\mathcal{K}}^2\right)^{\frac{1}{2}}
+ \left(\sum_{r=1}^R\|g_r\|_{\mathcal{G}}^2\right)^{\frac{1}{2}}\right]\\
\subjectto~&~p=\sum_{r=1}^R f_r g_r, ~f_r\in \mathcal{H}_{K},~g_r\in \mathcal{H}_{G}.\nonumber
\end{align}
\end{lemma}

\color{black}
Due to Lemma~\ref{le:sqrt}, the problem in \eqref{eq:regularization3} is reformulated, and $p$ can be learned via the regularization
\begin{subequations}\label{eq:QKG+QKGp}
\begin{equation}\label{eq:QKG}
Q(\mathcal{K},\mathcal{G}):=\min_{p\in\mathcal{P}} ~ Q(\mathcal{K},\mathcal{G},p)
\end{equation}
where 
\begin{align}\label{eq:QKGp}
Q(\mathcal{K},\mathcal{G},p) &:=\|\mathbf{Z}-\mathbf{P}\|_F^2\nonumber\\
&+ \mu\left(\sum_{r=1}^R \|f_r\|_{\mathcal{K}}^2\right)^{\frac{1}{2}}
+ \mu \left(\sum_{r=1}^R\|g_r\|_{\mathcal{G}}^2\right)^{\frac{1}{2}}.
\end{align}
\end{subequations}

\color{black}

\subsection{Multi-Kernel Learning}\label{subsec:mkl}
\color{black}
Solving the inference problem in \eqref{eq:QKG+QKGp} assumes that $\mu$ and the kernels $\mathcal{K}$ and $\mathcal{G}$ are known. The parameter $\mu$ is typically tuned via cross-validation~\cite{Hastie}. Choosing the appropriate kernels though is more challenging, as testified by the extensive research on \emph{multi-kernel learning}; see the reviews \cite{Gonen11}, \cite{AlRoLa12}. 
\color{black}

In this work, the multi-kernel learning approach of \cite{Mich05} is generalized to the function regularization in \eqref{eq:QKG+QKGp}. Specifically, two sets of kernel function choices, $\{K_l\}_{l=1}^L$ and $\{G_m\}_{m=1}^M$, are provided for nodes and time periods, respectively. \textcolor{black}{Numbers $L$ and $M$ are selected depending on the kernel choices and the computational resources available.} Consider the kernel spaces constructed as the convex hulls
\begin{subequations}\label{eq:convexhulls}
\begin{align}
\mathcal{K}&:=\{K=\sum_{l=1}^L \theta_l K_l,~\theta_l>0,~\sum_{l=1}^L\theta_l=1\}\label{eq:K}\\
\mathcal{G}&:=\{G=\sum_{m=1}^M \phi_m G_m,~\phi_m>0,~\sum_{m=1}^M\phi_m=1\}.\label{eq:G}
\end{align}
\end{subequations}
Optimizing the outcome of the regularization problem in \eqref{eq:QKG} over $\mathcal{K}$ and $\mathcal{G}$ provides a disciplined kernel design methodology. Since all $K_l$ and $G_m$ are predefined, minimizing \eqref{eq:QKG} over $\mathcal{K}$ and $\mathcal{G}$, reduces to minimizing $Q(\mathcal{K},\mathcal{G})$ over the weights $\{\theta_l\}$ and $\{\phi_m\}$. The following theorem, which is proved in Appendix~\ref{subsec:mkl_proof}, shows how the kernel learning part can be accomplished without even finding the optimal weights.

\begin{theorem}\label{th:mkl}
Consider the function space $\mathcal{P}$, the kernel spaces $\mathcal{K}$ and $\mathcal{G}$, and the functional $Q(\mathcal{K},\mathcal{G},p)$, defined in \eqref{eq:F}, \eqref{eq:convexhulls}, and \eqref{eq:QKGp}, respectively. Solving the regularization problem 
\begin{equation}\label{eq:double}
\min_{\mathcal{K},\mathcal{G}}\min_{p\in \mathcal{P}}~ Q(\mathcal{K},\mathcal{G},p)
\end{equation}
is equivalent to solving
\begin{equation}\label{eq:Pprime}
\min_{p\in \mathcal{P}'}  \|\mathbf{Z}-\mathbf{P}\|_F^2+ \mu\sum_{l=1}^L
\sqrt{\sum_{r=1}^R \|f_{lr}\|_{\mathcal{K}_l}^2}
+ \mu \sum_{m=1}^M\sqrt{\sum_{r=1}^R\|g_{mr}\|_{\mathcal{G}_m}^2}
\end{equation}
over $\mathcal{P}':=\left\{ p(n,t)=\sum_{r=1}^R f_r(n) g_r(t):f_r=\sum_{l=1}^L f_{lr},\right.$
$\left. f_{lr}\in \mathcal{H}_{\mathcal{K}_l},~g_r=\sum_{m=1}^M g_{mr},~g_{mr}\in \mathcal{H}_{\mathcal{G}_m}\right\}$, where $\{\mathcal{H}_{\mathcal{K}_l}\}$ and $\{\mathcal{H}_{\mathcal{G}_m}\}$ are the function spaces defined by the kernels $K_l$ and $G_m$, accordingly.
\end{theorem}

Theorem \ref{th:mkl} asserts that minimizing \eqref{eq:QKGp} over $f_r\in \mathcal{H}_{\mathcal{K}}$ and $g_r\in \mathcal{H}_{\mathcal{G}}$ boils down to the functional optimization in \eqref{eq:Pprime} where $f_r$ and $g_r$ are now simply decomposed as $\sum_{l=1}^L f_{lr}$ and $\sum_{m=1}^M g_{mr}$, respectively. Interestingly enough, the theorem also generalizes the multi-kernel learning results of \cite{Mich05} to the low-rank decomposition model of \eqref{eq:F}. After drawing some interesting connections in Section~\ref{subsec:connections}, the functional inference in \eqref{eq:Pprime} is transformed to a matrix minimization problem in Section~\ref{sec:matrix}.

\color{black}
\subsection{Interesting Connections}\label{subsec:connections}
Observe that when $\mathcal{N}$ and $\mathcal{T}$ are Euclidean spaces, $K(n,n')=\delta(n-n')$ and $G(t,t')=\delta(t-t')$ where $\delta(\cdot)$ is the Kronecker delta function, then $\mathcal{P}$ in \eqref{eq:F} is the space of matrices $\mathbf{P}\in\mathbb{R}^{|\mathcal{N}|\times |\mathcal{T}|}$ having $p(n,t)$ as their $(n,t)$-th entry. In this case, $\|p\|_*$ is simply the nuclear norm $\|\mathbf{P}\|_*$ of matrix $\mathbf{P}$, i.e., the sum of its singular values; $R=\rank(\mathbf{P})$; and \eqref{eq:tracenorm} becomes~\cite{Aber09}, \cite{BaGia13},
\begin{align}\label{eq:tracenorm_matrix}
\|\mathbf{P}\|_*=\min_{\mathbf{F},\mathbf{G}} ~&~\tfrac{1}{2}\left(\|\mathbf{F}\|_F^2+\|\mathbf{G}\|_F^2\right)\\ 
\subjectto~&~\mathbf{P}=\mathbf{F}\mathbf{G}^T, ~\mathbf{F}\in \mathbb{R}^{|\mathcal{N}|\times R},~\mathbf{G}\in\mathbb{R}^{|\mathcal{T}|\times R}.\nonumber
\end{align}
The alternative representation of $\|\mathbf{P}\|_*$ in \eqref{eq:tracenorm_matrix} has been extensively used in nuclear norm minimization \cite{SrSh05}, \cite{ReFaPa10}, \cite{MaMaGia13}. Interestingly, the matrix analogue of Lemma~\ref{le:sqrt} reads:
\begin{corollary}\label{co:sqrt}
For $\mathbf{P}\in\mathbb{R}^{N\times T}$ with $\rank(\mathbf{P})=R$, it holds
\begin{align}\label{eq:tracenorm_matrix_sqrt}
\|\mathbf{P}\|_*^{1/2}=\min_{\mathbf{F},\mathbf{G}} ~&~\tfrac{1}{2}\left(\|\mathbf{F}\|_F+\|\mathbf{G}\|_F\right)\\ 
\subjectto~&~\mathbf{P}=\mathbf{F}\mathbf{G}^T, ~\mathbf{F}\in \mathbb{R}^{N\times R},~\mathbf{G}\in\mathbb{R}^{T\times R}.\nonumber
\end{align}
\end{corollary}

Matrix completion aims at recovering a low-rank matrix $\mathbf{P}$ given noisy measurements for a few of its entries~\cite{Fazel02}. It can be derived from \eqref{eq:regularization2.5} after replacing $\|p\|_*$ by $\|\mathbf{P}\|_*$ [or \eqref{eq:tracenorm_matrix}], and $\|\mathbf{Z}-\mathbf{P}\|_F^2$ by $\|(\mathbf{Z}-\mathbf{P}) \odot\mathbf{\Delta}\|_F^2$, where $\odot$ denotes element-wise multiplication and $\mathbf{\Delta}$ is a binary matrix having zeros on the missing entries. The premise is that $\mathbf{P}$ could be recovered due to its low-rank property. But recovery is impossible when entire columns or rows are missing.

For generic \emph{yet fixed} kernels $K(n,n')$ and $G(t,t')$, low-rank kernel-based models could be similarly derived as special cases of \eqref{eq:regularization2.5}; see e.g., \cite{Aber09}, \cite{BaGia13}. Using kernel functions other than the Kronecker delta, enables not only recovering the missing entries, but extrapolating to unseen columns and rows as well. Different from matrix completion and low-rank kernel-based inference, our regularization in \eqref{eq:Pprime} targets to jointly learn a low-rank $p(n,t)$, together with kernels $K$ and $G$.

\color{black}
\section{Matrix Optimization}\label{sec:matrix}
The next goal is to map the functional optimization of \eqref{eq:Pprime} to a vector minimization by resorting to the Representer's Theorem \cite{Hastie}. Observe that minimizing \eqref{eq:Pprime} over a specific $f_{lr}$ is actually a functional minimization regularized by $(\|f_{lr}\|_{\mathcal{K}_l}^2+c_{lr})^{1/2}$ for some constant $c_{lr}\geq 0$. Since the regularization is an increasing function of $\|f_{lr}\|_{\mathcal{K}_l}^2$, Representer's Theorem applies readily~\cite{Hastie}, \cite{Argy09}. 

Each one of the $LR$ functions $f_{lr}$ minimizing \eqref{eq:Pprime} can be expressed as a linear combination of the associated kernel $K_l$ evaluated over the $N$ training examples involved, that is
\begin{equation}\label{eq:flr_RT}
f_{lr}(n)=\sum_{n'=1}^N K_l(n,n')\beta_{lr,n'}.
\end{equation}
Upon concatenating the unknown expansion coefficients and the function values into $\boldsymbol{\beta}_{lr}:=[\beta_{lr,1}~\cdots~\beta_{lr,N}]^{\top}$ and $\mathbf{f}_{lr}:=[f_{lr}(1)~\cdots~f_{lr}(N) ]^{\top}$, respectively, it holds that
\begin{equation}\label{eq:flr}
\mathbf{f}_{lr}= \mathbf{K}_l\boldsymbol{\beta}_{lr}
\end{equation}
where $\mathbf{K}_l\in \mathbb{S}^{N}_{++}$ is the node kernel matrix whose $(n,n')$-th entry is $K_l(n,n')$. Using \eqref{eq:flr} and accounting for the decomposition $f_r=\sum_{l=1}^L f_{lr}$ dictated by \eqref{eq:Pprime}, the vector collecting the values $\{f_r(n)\}_{n=1}^N$ is compactly written as
\begin{equation}\label{eq:fr}
\mathbf{f}_{r}= \sum_{l=1}^L \mathbf{K}_l\boldsymbol{\beta}_{lr}.
\end{equation}

Likewise, each $g_{mr}$ minimizing \eqref{eq:Pprime} admits the expansion
\begin{equation}\label{eq:gmr_RT}
g_{mr}(t)= \sum_{t'=1}^T G_m(t,t') \gamma_{mr,t'}
\end{equation}
for all $t$. Similar to \eqref{eq:flr}, the vector of function values $\mathbf{g}_{mr}:=[g_{mr}(1)~\ldots~g_{mr}(T)]^{\top}$ is expressed in terms of the time kernel matrix $\mathbf{G}_m\in\mathbb{S}_{++}^T$ as  
\begin{equation}\label{eq:gmr}
\mathbf{g}_{mr}= \mathbf{G}_m \boldsymbol{\gamma}_{mr}
\end{equation}
where $\boldsymbol{\gamma}_{mr}:=[\gamma_{mr,1}~\ldots~\gamma_{mr,T}]^{\top}$. Due to the decomposition $g_r=\sum_{m=1}^M g_{mr}$ in \eqref{eq:Pprime}, the vector containing $\{g_r(t)\}_{t=1}^{T}$ is provided by [cf.~\eqref{eq:fr}]
\begin{equation}\label{eq:gr}
\mathbf{g}_{r}= \sum_{m=1}^M \mathbf{G}_m \boldsymbol{\gamma}_{mr}.
\end{equation}

So far, the functions $\{f_r(n),g_r(t)\}_{r=1}^R$ minimizing \eqref{eq:Pprime} have been expressed in terms of $\boldsymbol{\beta}_{lr}$'s and $\boldsymbol{\gamma}_{mr}$'s, thus enabling one to transform \eqref{eq:Pprime} to a minimization problem over the unknown coefficients. 

Regarding the price matrix $\mathbf{P}$, the low-rank model $p(n,t)=\sum_{r=1}^R f_r(n) g_r(t)$ implies that
\begin{equation}\label{eq:Pfr}
\mathbf{P} = \sum_{r=1}^R\mathbf{f}_r\mathbf{g}_r^{\top}.
\end{equation}
Plugging \eqref{eq:fr} and \eqref{eq:gr} into \eqref{eq:Pfr}, yields 
\begin{equation}\label{eq:P}
\mathbf{P} = \sum_{l=1}^L \sum_{m=1}^M \mathbf{K}_l \mathbf{B}_l \mathbf{\Gamma}_m^{\top}\mathbf{G}_m
\end{equation}
where $\mathbf{B}_l:=[\boldsymbol{\beta}_{l1}~\cdots~\boldsymbol{\beta}_{lR}]\in \mathbb{R}^{N\times R}$ and $\mathbf{\Gamma}_m:=[\boldsymbol{\gamma}_{m1}~\cdots~\boldsymbol{\gamma}_{mR}]\in \mathbb{R}^{T\times R}$ for all $l$ and $m$.

Consider now the regularization terms in \eqref{eq:Pprime}. Due to \eqref{eq:flr_RT} and \eqref{eq:gmr_RT}, the associated norms can be written as $\|f_{lr}\|_{\mathcal{K}_l}^2=\boldsymbol{\beta}_{lr}^{\top} \mathbf{K}_l\boldsymbol{\beta}_{lr}$ and $\|g_{mr}\|_{\mathcal{G}_m}^2=\boldsymbol{\gamma}_{mr}^{\top} \mathbf{G}_m\boldsymbol{\gamma}_{mr}$~[cf.~\eqref{eq:family}-\eqref{eq:regularization2}]. Using the properties of the trace operator, it can be shown that
\begin{subequations}\label{eq:traces}
\begin{align}
\sum_{r=1}^R \|f_{lr}\|_{\mathcal{K}_l}^2 &=\trace(\mathbf{B}_l^{\top}{\mathbf{K}_l}\mathbf{B}_l)\label{eq:traceBl}
\\ \sum_{r=1}^R \|g_{mr}\|_{\mathcal{G}_m}^2 &=\trace(\mathbf{\Gamma}_m^{\top}{\mathbf{G}_m}\mathbf{\Gamma}_m).\label{eq:traceGammam}
\end{align}
\end{subequations}
The right-hand sides in \eqref{eq:traces} can be identified as the norms $\|\mathbf{B}_l\|_{\mathbf{K}_l}^2:=\trace(\mathbf{B}_l^{\top}{\mathbf{K}_l}\mathbf{B}_l)$ and $\|\mathbf{\Gamma}_m \|_{\mathbf{G}_m}^2:= \trace(\mathbf{\Gamma}_m^{\top}{\mathbf{G}_m}\mathbf{\Gamma}_m)$. By using \eqref{eq:P}-\eqref{eq:traces}, the functional optimization in \eqref{eq:Pprime} can be compactly expressed as the matrix optimization problem
\begin{align}
\min_{\mathbf{P},\{\mathbf{B}_l\},\{\mathbf{\Gamma}_m\}} &~\|\mathbf{Z}-\mathbf{P}\|_F^2+ \mu\sum_{l=1}^L\|\mathbf{B}_l\|_{\mathbf{K}_l}
+ \mu \sum_{m=1}^M\|\mathbf{\Gamma}_m\|_{\mathbf{G}_m}\nonumber\\
\subjectto~&~ \mathbf{P} =\sum_{l=1}^L \sum_{m=1}^M \mathbf{K}_l \mathbf{B}_l \mathbf{\Gamma}_m^{\top}\mathbf{G}_m. \label{eq:problem}
\end{align}
Solving \eqref{eq:problem} faces two challenges. Even though optimizing separately over $\{\mathbf{B}_l\}$ or $\{\mathbf{\Gamma}_m\}$ entails a convex cost, the joint minimization is non-convex. Secondly, solving \eqref{eq:problem} involves multiple high-dimensional matrices, which raises computational concerns. The algorithm developed in the next section scales well with the problem dimensions, and converges to a stationary point of \eqref{eq:problem}.

\emph{Price Forecasting:} Having found all $\hat{\mathbf{B}}_l$ and $\hat{\mathbf{\Gamma}}_m$, the electricity prices over the training period can be reconstructed via \eqref{eq:Pfr}. Of course, the ultimate learning goal is inferring future prices. Based on the modeling approach in Section~\ref{sec:approach}, the price $p(n_0,t_0)$ for an unseen pair $(n_0,t_0)$ can be predicted simply as
\begin{equation}\label{eq:prediction}
\hat{p}(n_0,t_0)=\sum_{r=1}^R \sum_{l=1}^L\sum_{m=1}^M \hat{f}_{lr}(n_0)\hat{g}_{mr}(t_0)
\end{equation}
where $\hat{f}_{lr}(n_0)=\sum_{n=1}^N K_l(n_0,n)\hat{\beta}_{lr,n}$ and $\hat{g}_{mr}(t_0)=\sum_{t=1}^{T} G_m(t_0,t) \hat{\gamma}_{mr,t}$ [cf.~\eqref{eq:flr_RT}, \eqref{eq:gmr_RT}]. In essence, extrapolation to $(n_0,t_0)$ is viable conditioned on availability of the kernel values involved. 

\color{black}

If network-wide forecasts are needed over a future interval $\mathcal{T}'$ and over the node set $\mathcal{N}'$, the predicted values can be stored in the $|\mathcal{N}'|\times |\mathcal{T}'|$ matrix $\hat{\mathbf{P}}'$. According to \eqref{eq:prediction}, matrix $\hat{\mathbf{P}}'$ is compactly expressed as 
\begin{equation}\label{eq:prediction_mat}
\hat{\mathbf{P}}'=\sum_{m=1}^M \sum_{l=1}^L \mathbf{K}_l'\hat{\mathbf{B}}_l\hat{\mathbf{\Gamma}}_m^{\top}\mathbf{G}_m'
\end{equation}
where $\mathbf{K}_l'\in \mathbb{R}^{N\times |\mathcal{N}'|}$ and $\mathbf{G}_m'\in\mathbb{R}^{T\times |\mathcal{T}'|}$ are the kernel matrices between the training and the forecast points, i.e., having entries $[\mathbf{K}_l']_{n,n'}=K_l(n,n')$ and $[\mathbf{G}_m']_{t,t'}=G_m(t,t')$. Important remarks are now in order.

\begin{remark} Price forecasts are not confined to future $t_0$'s (essentially unseen feature vectors $\mathbf{x}_{t_0}$'s); they can be issued even for a new node $n_0\notin \mathcal{N}$. This is an important feature when dealing with electricity markets having seasonal pricing models. For example, MISO updates its commercial grid quarterly by adding, removing, merging, and redefining CPNs, to accommodate transmission grid updates and market participants leaving or entering the market.
\end{remark}

\color{black}
\begin{remark}
In addition to extrapolation (prediction), the proposed approach is general enough to encompass imputation of missing entries. Similar to matrix completion [cf.~Section~\ref{subsec:connections}], that would be possible upon substituting $\|\mathbf{Z}-\mathbf{P}\|_F^2$ in \eqref{eq:problem} by $\|(\mathbf{Z}-\mathbf{P}) \odot\mathbf{\Delta}\|_F^2$.
\end{remark}

\begin{remark} As justified in Sec.~\ref{sec:bcd}, \eqref{eq:problem} promotes \emph{block-sparse solutions}. In particular, some of the $\{\hat{\mathbf{B}}_l\}_{l=1}^L$ and $\{\hat{\mathbf{\Gamma}}_m\}_{m=1}^M$ may be driven to zero. The latter indicates that the corresponding $K_l$ or $G_m$ are not influential in price clearing. Since experimentation with kernels defined over different feature subsets can be highly interpretative, the proposed approach becomes a systematic prediction and kernel selection tool.
\end{remark}

\section{Block-Coordinate Descent Algorithm}\label{sec:bcd}
A block-coordinate descent (BCD) algorithm is developed here for solving \eqref{eq:problem}. According to the BCD methodology, the initial optimization variable is partitioned into blocks. Per block minimizations having the remaining variables fixed are then iterated cyclically over blocks.

Solving \eqref{eq:problem} in particular, variable blocks are selected in the order $\{\mathbf{B}_1,\ldots,\mathbf{B}_L,\mathbf{\Gamma}_1,\ldots,\mathbf{\Gamma}_M\}$. The per block minimizations involved are detailed next. Consider minimizing \eqref{eq:problem} over a specific $\mathbf{B}_l$, while all other variables are maintained to their most recent values $\{\hat{\mathbf{B}}_{l'}\}_{l'\neq l}$ and $\{\hat{\mathbf{\Gamma}}_m\}_{m=1}^M$. Upon rearranging terms in \eqref{eq:problem}, block $\mathbf{B}_l$ can be updated as
\begin{equation}\label{eq:problemBl}
\hat{\mathbf{B}}_l = \arg\min_{\mathbf{B}_l} ~ \|\mathbf{Z}_l^B-\mathbf{K}_l \mathbf{B}_l \mathbf{H}^{\top}\|_F^2+ \mu\|\mathbf{B}_l\|_{\mathbf{K}_l}
\end{equation}
where $\mathbf{H}:=\sum_{m=1}^M \mathbf{G}_m\hat{\mathbf{\Gamma}}_m$ is the contribution of all $\hat{\mathbf{\Gamma}}_m$, and $\mathbf{Z}_l^B:=\mathbf{Z} - \sum_{l'\neq l} \mathbf{K}_{l'}\hat{\mathbf{B}}_{l'}\mathbf{H}^{\top}$. 

Similarly, updating a particular $\mathbf{\Gamma}_m$ entails finding
\begin{equation}\label{eq:problemGammam}
\hat{\mathbf{\Gamma}}_m = \arg\min_{\mathbf{\Gamma}_m} ~ \|\mathbf{Z}_m^{\Gamma}-\mathbf{F} \mathbf{\Gamma}_m^{\top} \mathbf{G}_m\|_F^2+ \mu \|\mathbf{\Gamma}_m\|_{\mathbf{G}_m}
\end{equation}
where $\mathbf{F}:=\sum_{l=1}^L \mathbf{K}_l\hat{\mathbf{B}}_l$ is the contribution of all $\hat{\mathbf{B}}_l$, and $\mathbf{Z}_m^{\Gamma}:=\mathbf{Z} -\sum_{m'\neq m}  \mathbf{F}\mathbf{\Gamma}_{m'}^{\top}\mathbf{G}_{m'}$.

Problems \eqref{eq:problemBl} and \eqref{eq:problemGammam} are convex, yet not differentiable, and exhibit the same canonical form. This form can be efficiently solved according to the following lemma that is proved in Appendix~\ref{subsec:canonical_proof}. 

\begin{lemma}\label{le:canonical}
Let $\mathbf{A}\in \mathbb{R}^{d_1\times d_3}$, $\mathbf{B}\in \mathbb{S}^{d_1}_{++}$, $\mathbf{C}\in\mathbb{R}^{d_3\times d_2}$, and $\mu>0$. The convex optimization problem
\begin{equation}\label{eq:canonical}
\min_{\mathbf{X}} ~ \|\mathbf{A}-\mathbf{B}\mathbf{X}\mathbf{C}^{\top}\|_F^2 + \mu \|\mathbf{X}\|_{\mathbf{B}}
\end{equation}
has a unique minimizer $\hat{\mathbf{X}}$ provided by the solution of
\begin{equation}\label{eq:sylvester}
\mathbf{B}\hat{\mathbf{X}}\mathbf{C}^{\top}\mathbf{C} + \frac{\mu^2}{4\hat{w}} \hat{\mathbf{X}} = \mathbf{A}\mathbf{C}
\end{equation}
if $\|\mathbf{B}^{1/2}\mathbf{A}\mathbf{C}\|_F>\mu/2$; or, $\hat{\mathbf{X}} = \mathbf{0}$, otherwise. The scalar $\hat{w}>0$ in \eqref{eq:sylvester} is the minimizer of the convex problem
\begin{equation}\label{eq:univariate}
\hat{w}:=\arg\min_{w\geq 0}~ w - \sum_{i=1}^{d_1} \sum_{j=1}^{d_2} \frac{[\mathbf{W}]_{ij}^2\lambda_i\mu_j w}{\lambda_i\mu_j w + \mu^2/4}
\end{equation}
where $\mathbf{W}:=\mathbf{U}_B^{\top}\mathbf{A}\mathbf{U}_C$; $(\mathbf{U}_B,\{\lambda_i\}_{i=1}^{d_1})$ are the eigenpairs of $\mathbf{B}$; and $(\mathbf{U}_C,\{\mu_j\}_{j=1}^{d_2})$ the non-zero eigenpairs of $\mathbf{C}\mathbf{C}^{\top}$.
\end{lemma}

Lemma~\ref{le:canonical} provides valuable insights for solving \eqref{eq:canonical}. It reveals that by simply calculating $\|\mathbf{B}^{1/2}\mathbf{A}\mathbf{C}\|_F$, the sought $\hat{\mathbf{X}}$ may be directly set to zero. Hence, \eqref{eq:canonical} admits block-zero minimizers depending on the value of $\mu$. This property critically implies that some of the $\{\hat{\mathbf{B}}_l\}$ and $\{\hat{\mathbf{\Gamma}}_m\}$ minimizing \eqref{eq:problem} will be zero, thus, effecting kernel selection. 

\begin{algorithm}[t]
\caption{Minimizing the canonical form \eqref{eq:canonical}}\label{alg:canonical}
\begin{algorithmic}[1]
\renewcommand{\algorithmicrequire}{\textbf{Input:}}
\renewcommand{\algorithmicensure}{\textbf{Output:}}
\Function{SolveCanonical}{$\mathbf{A}$,$\mathbf{B}$,$\mathbf{C}$,$\mu$} 
\If{$\|\mathbf{B}^{1/2}\mathbf{A}\mathbf{C}\|_F\leq \mu/2$} 
$\hat{\mathbf{X}}=\mathbf{0}$ 
\Else
\State $\left(\mathbf{U}_B,\{\lambda_i\}\right)$ = \Call{EigenDecomposition}{$\mathbf{B}$}
\State $\left(\mathbf{U}_C,\{\mu_j\}\right)$ = \Call{EigenDecomposition}{$\mathbf{C}\mathbf{C}^{\top}$}
\State Define $\mathbf{W}=\mathbf{U}_B^{\top}\mathbf{A}\mathbf{U}_C$
\State Initialize $w^0=0$ and $t=0$
\Repeat 
\State Evaluate $s'(w^t)$ via \eqref{eq:derivative}
\State Update $w^{t+1}=\max\left\{0,w^t - c\cdot s'(w^t)\right\}$
\State $t=t+1$
\Until{$\left|s(w^{t}) -s(w^{t-1})\right|<\epsilon_{\textrm{c}}$}
\State Set $\hat{w}=w^{t}$
\State Obtain $\hat{\mathbf{X}}$ by solving the Sylvester equation \eqref{eq:sylvester}
\EndIf
\EndFunction
\end{algorithmic}
\end{algorithm}

Back to Lemma~\ref{le:canonical}, if $\|\mathbf{B}^{1/2}\mathbf{A}\mathbf{C}\|_F> \mu/2$, a non-zero solution emerges. The univariate optimization in \eqref{eq:univariate} and the linear matrix equations in \eqref{eq:sylvester} can be efficiently tackled as described next. First, the constrained convex problem in \eqref{eq:univariate} can be solved by a projected gradient algorithm. If $s(w)$ denotes the cost function in \eqref{eq:univariate}, its derivative is
\begin{equation}\label{eq:derivative}
s'(w)=1-\sum_{i=1}^{d_1} \sum_{j=1}^{d_2} \frac{\mu^2[\mathbf{W}]_{ij}^2\lambda_i\mu_j }{4\left(\lambda_i\mu_j w + \mu^2/4 \right)^2}.
\end{equation}
\textcolor{black}{The iterates $w^{t+1}=\max\left\{0,w^t - c\cdot s'(w^t)\right\}$ are guaranteed to converge to the global minimum $\hat{w}$ for a sufficiently small step size $c>0$; see \cite{Be99} for details. Each iterate costs $\mathcal{O}(d_1d_2)$ operations.} 

\textcolor{black}{Secondly, concerning \eqref{eq:sylvester}, it can be rewritten as a Sylvester equation as advocated also in \cite{KeVeLiGia13}, \cite{SiLoMi12}. Hence, $\hat{\mathbf{X}}$ can be found in $\mathcal{O}(d_1^3 + d_2^3)$ numerical operations using the Bartels-Stewart algorithm \cite[Alg.~7.6.2]{GolubVanLoan}, instead of the $\mathcal{O}(d_1^3d_2^3)$ complexity of a generic linear system solver. The steps for solving the canonical problem \eqref{eq:canonical} have been tabulated as Alg.~\ref{alg:canonical}, while its overall worst-cast complexity is $\mathcal{O}(d_1^3+d_2d_3^2)$.}

\begin{algorithm}[t]
\caption{BCD algorithm for solving \eqref{eq:problem}}\label{alg:BCD}
\begin{algorithmic}[1]
\renewcommand{\algorithmicrequire}{\textbf{Input:}}
\renewcommand{\algorithmicensure}{\textbf{Output:}}
\Require $\mathbf{Z}$, $\{\mathbf{K}_l\}_{l=1}^L$, $\{\mathbf{G}_m\}_{m=1}^M$, $R$, $\mu$
\State Randomly initialize $\{\hat{\mathbf{B}}_l\}_{l=1}^L$ and $\{\hat{\mathbf{\Gamma}}_m\}_{m=1}^M$
\State Compute $\mathbf{F}=\sum_{l=1}^L \mathbf{K}_l\hat{\mathbf{B}}_l$ and $\mathbf{H}=\sum_{m=1}^M \mathbf{G}_m\hat{\mathbf{\Gamma}}_m$
\State Store $\{\hat{\mathbf{B}}_l^{\textrm{old}}=\hat{\mathbf{B}}_l\}_{l=1}^L$ and $\{\hat{\mathbf{\Gamma}}_m^{\textrm{old}}=\hat{\mathbf{\Gamma}}_m\}_{m=1}^M$
\Repeat
\For{$l = 1 \to L$} 
\State Update $\mathbf{F} = \mathbf{F} - \mathbf{K}_l\hat{\mathbf{B}}_l$
\State Define $\mathbf{Z}_l^{B} = \mathbf{Z} - \mathbf{F}\mathbf{H}^{\top}$
\State $\hat{\mathbf{B}}_l=$ \Call{SolveCanonical}{$\mathbf{Z}_l^{B}$,$\mathbf{K}_l$,$\mathbf{H}$,$\mu$}
\State Update $\mathbf{F} = \mathbf{F} + \mathbf{K}_l\hat{\mathbf{B}}_l$
\EndFor
\For{$m = 1 \to M$}
\State Update $\mathbf{H} = \mathbf{H} - \mathbf{G}_m\hat{\mathbf{\Gamma}}_m$
\State Define $\mathbf{Z}_m^{\Gamma} = \mathbf{Z} - \mathbf{F}\mathbf{H}^{\top}$
\State $\hat{\mathbf{\Gamma}}_m=$ \Call{SolveCanonical}{$(\mathbf{Z}_m^{\Gamma})^{\top}$,$\mathbf{G}_m$,$\mathbf{F}$,$\mu$}
\State Update $\mathbf{H} = \mathbf{H} + \mathbf{G}_m\hat{\mathbf{\Gamma}}_m$
\EndFor
\Until{$\left|\frac{f(\{\hat{\mathbf{B}}_l\},\{\hat{\mathbf{\Gamma}}_m\})}{f(\{\hat{\mathbf{B}}_l^{\textrm{old}}\},\{\hat{\mathbf{\Gamma}}_m^{\textrm{old}}\})}-1\right|<\epsilon_{\textrm{BCD}}:$ $f(\cdot)$ is the cost in \eqref{eq:problem}}
\Ensure $\{\hat{\mathbf{B}}_l\}_{l=1}^L$, $\{\hat{\mathbf{\Gamma}}_m\}_{m=1}^M$
\end{algorithmic}
\end{algorithm}

\color{black}
Proceeding with the BCD steps \eqref{eq:problemBl} and \eqref{eq:problemGammam}, those can be efficiently performed after carefully updating $\mathbf{H}$ and $\mathbf{F}$. The final steps for solving \eqref{eq:problem} are listed as Alg.~\ref{alg:BCD}. Due to the separability of the non-differentiable cost over the chosen variable blocks, the BCD algorithm is guaranteed to converge to a stationary point of \eqref{eq:problem}~\cite{Tseng01}. The BCD iterates are terminated when the relative cost value error becomes smaller than some threshold $\epsilon_{\textrm{BCD}}=10^{-3}$. The eigendecomposition of all kernel matrices can be computed once. Algorithm~\ref{alg:BCD} has a complexity of $\mathcal{O}\left( L(N^3+RT^2)+M(T^3+RN^2)\right)$ per iteration. In the numerical experiments of Section~\ref{sec:simulations}, and depending on the value of $\mu$, 5-15 BCD iterations were sufficient.

\color{black}
\section{Numerical Tests}\label{sec:simulations}
The derived low-rank multi-kernel learning approach was tested using real data from the Midwest ISO (MISO) electricity market. Day-ahead hourly LMPs were collected across $N=1,732$ nodes for the period June 1 to August 31, 2012, yielding a total of 92 days or 2,208 hours. 

\begin{figure}[t]
\centering
\includegraphics[width=1.0\linewidth]{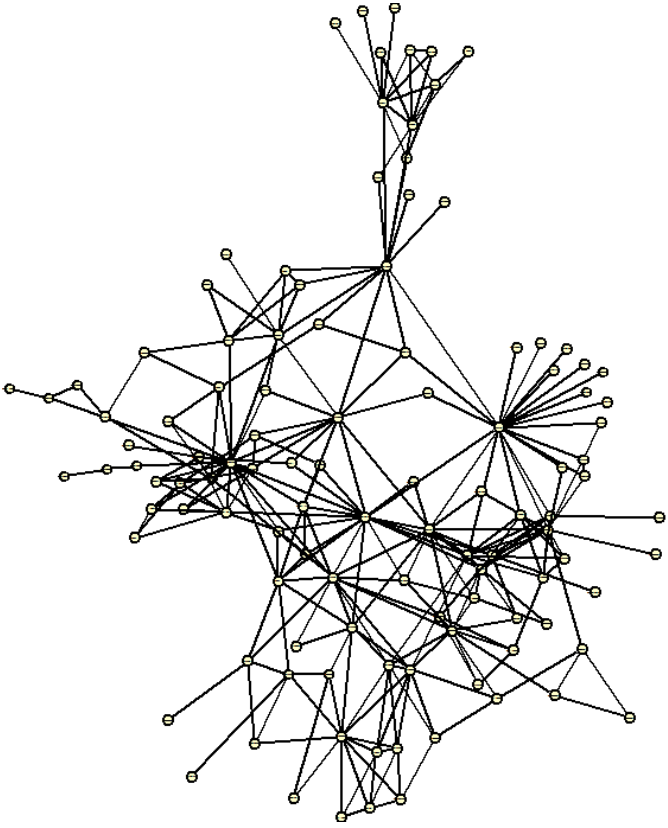}
\caption{Graph of the LBAs involved in the MISO market.}
\label{fig:lba}
\end{figure}

A pool of $K=5$ nodal and $L=5$ time kernels was selected as detailed next. Starting with the nodal ones, when learning over a graph, the corresponding graph Laplacian matrix is oftentimes used to design meaningful kernels~\cite{Kolaczyk}. CPNs are considered here as vertices of a similarity graph, connected with edges having non-negative weights proportional to the similarity between incident CPNs. Nonetheless, lacking any other type of geographical or electrical distance, the local balancing authority (LBA) each CPN belongs to was adopted here as a topology surrogate. The presumption is that nodes of the same LBA experience similar prices. Further, nodes controlled by neighboring authorities are expected to have prices correlated more than nodes under non-adjacent ones. The connectivity graph of 131 LBAs involved in MISO was constructed based on publicly available data found on MISO's website; cf. Fig.~\ref{fig:lba}.

Kernel matrices $\mathbf{K}_1,~\mathbf{K}_2\in\mathbb{S}_{++}^N$ were built based on this LBA connectivity graph as follows. Edges between CPNs of the same LBA were assigned unit weights; edges across CPNs from different LBAs received weight $0.5$; and all other edges were set to zero. If weight values are stored in the adjacency matrix $\mathbf{A}_{\mathcal{N}}$, the normalized Laplacian matrix of a graph is defined as $\mathbf{L}_{\mathcal{N}}:=\mathbf{I}_N - \mathbf{D}_{\mathcal{N}}^{-1/2} \mathbf{A}_{\mathcal{N}}\mathbf{D}_{\mathcal{N}}^{-1/2}$, where $\mathbf{D}_{\mathcal{N}}$ is a diagonal matrix with diagonal entries the row sums of $\mathbf{A}_{\mathcal{N}}$~\cite{Kolaczyk}. Then, $\mathbf{K}_1$ was selected as the regularized Laplacian $\mathbf{K}_1 :=\left(\mathbf{L}_{\mathcal{N}} + \mathbf{I}_N \right)^{-1}$, and $\mathbf{K}_2$ as the diffusion Laplacian $\mathbf{K}_2 :=\exp(-3\mathbf{L}_{\mathcal{N}})$~\cite{SmoKon03}.

Kernel $\mathbf{K}_3$ utilized information that could be infered from CPN names. Specifically, the prefix of every CPN name in MISO denotes its LBA, while some CPNs have similar names. For example, nodes ALTE.COLUMBAL1 and ALTE.COLUMBAL2 belong to the LBA named ALTE, and they are assumed to be geographically colocated. Every CPN is classified in the MISO market as generator, load, interface, or hub. The LBA, the name similarity, and the CPN type, were all used as categorical features by a Gaussian kernel whose bandwidth was fixed to the median of all pairwise squared Euclidean distances.

To capture potential independence across nodes, kernel $\mathbf{K}_4$ was chosen to be the identity matrix. The last nodal kernel $\mathbf{K}_5$ was selected as the covariance matrix of market prices empirically estimated using historical data.

\color{black}
Regarding temporal kernels $\{\mathbf{G}_m\}_{m=1}^5$, the following publicly available features were used:
\begin{enumerate}
\item Yesterday's day-ahead LMPs for the same hour.
\item Load forecasts for the north, south, and central regions of MISO footprint.
\item Generation capacity outage publicized by MISO.
\item Market-wide wind energy generation forecast issued by MISO.
\item Hourly temperature and humidity in major cities across the MISO footprint (Bismarck, Des Moines, Detroit, Kansas City, Milwaukee, Minneapolis). Instead of predicted values, the actual values recorded by the National Oceanic and Atmospheric Administration (NOOA) were used.
\item Binary encoded categorical features of hour of the day, day of the week, and a holiday indicator.
\end{enumerate}

For all but the categorical features, their one-hour delayed and one-hour advanced values were also considered. For example, the market forecast for 3pm depended on temperature forecasts for 2pm, 3pm, and 4pm. The reason was to model wind power and weather volatility, as well as time coupling across hours introduced by unit commitment as exemplified next: Having a high temperature forecast for 4pm increases the load demand at 4pm and 5pm. Additionally, industrial consumers aware of the weather forecast may start their cooling systems at 3pm or even earlier to save money and achieve space cooling by 4pm. Secondly, weather forecasts are characterized by delay uncertainties: a 24-hour ahead weather model predicts quite accurately that high winds or a cold wave will be coming say in the afternoon, yet the exact hour is not precisely known. Third, many generation units have physical constraints: e.g., once they are started, they should remain on for at least a specific number of hours; see e.g., \cite{SPM2013}. Such constraints introduce time-coupling across power generation ranges and hence prices.

\color{black}

Temporal kernels $\mathbf{G}_1$ to $\mathbf{G}_3$ were designed by plugging the aforementioned features into Gaussian kernels of bandwidths 1, 430 (the median of all pairwise Euclidean feature distances), and $10^4$, respectively. Kernel $\mathbf{G}_4$ was the Gaussian kernel obtained from all but the time-shifted features, and with its bandwidth set to the median of all pairwise Euclidean feature distances. Finally, $\mathbf{G}_5$ was selected as the linear kernel. As a standard preprocessing step, both nodal and temporal features were centered and standardized, while all $\mathbf{K}_l$'s and $\mathbf{G}_m$'s were normalized to unit diagonal elements.

Market data are cyclo-stationary: the market-wide price mean fluctuates hourly, yet with a period of one day. To cope with cyclo-stationarity, market prices in $\mathbf{Z}$ were centered upon subtracting the per-hour sample mean. The developed predictor will hence forecast the mean-compensated prices, and not the actual ones. It is important to mention though that usually the \emph{price differences} across CPNs, rather than absolute nodal prices, are of interest. This is because bilateral transactions and power transfer contracts depend on exactly such nodal differentials~\cite{Oren06}. In such cases, our price forecasts can be readily used. Otherwise, a simple market-wide price mean predictor could be easily trained. 

Several factors not captured by the publicly available features used here (e.g., transmission and generation outages) can severely affect the market. Due to this source of non-stationarity, the designed day-ahead predictors depend on market data only from the previous week. Hence, the dimension $T$ of $\mathbf{Z}$ and $\mathbf{P}$ in \eqref{eq:problem} is 168 (hours). 

Tuning the regularization parameter $\mu$ was based on market data from the first 14 days. The causal nature of the market did not allow shuffling data across time, as it is typically done in cross-validation. Instead, days 1-7 were used to predict day 8, days 2-8 for day 9, and the process was repeated up to day 14. The value of $\mu$ attaining the lowest prediction root mean square error (RMSE) over a grid of values was fixed when predicting all the remaining 78 evaluation days.

\begin{figure}
\centering
\subfigure[Singular values for actual price matrices $\mathbf{Z}$.]{
\includegraphics[width=3.3in]{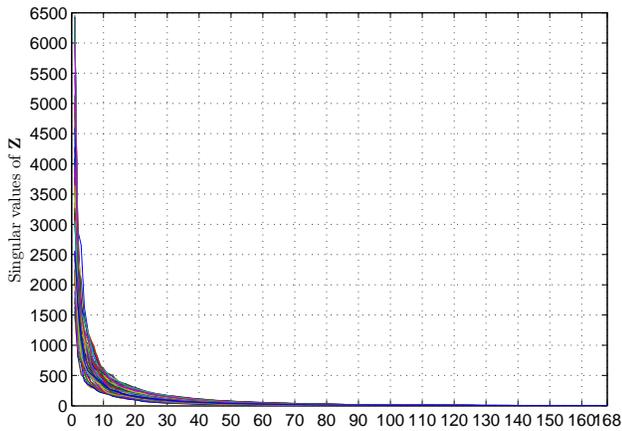}
\label{subfig:Zsvd}}\\
\subfigure[Singular values for predicted price matrices $\hat{\mathbf{P}}$.]{
\includegraphics[width=3.3in]{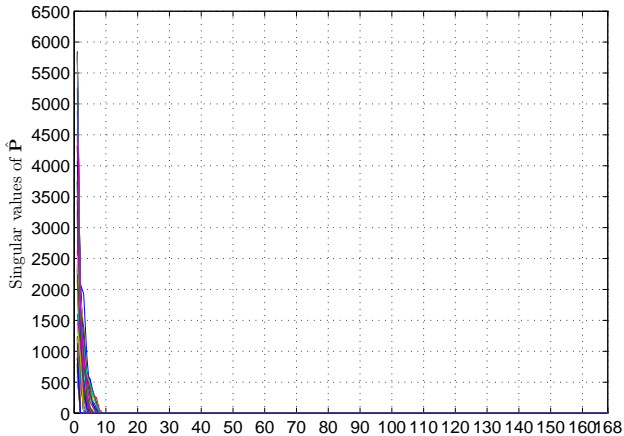}
\label{subfig:Psvd}}
\caption{Empirical distribution for the sorted singular values of price matrices: (a) for actual price matrices $\mathbf{Z}\in\mathbb{R}^{1732\times 168}$; and (b) for predicted price matrices $\hat{\mathbf{P}}$ as obtained by \eqref{eq:problem} for $R=20$.}
\label{fig:svd}
\end{figure}

\color{black}
Figure~\ref{subfig:Zsvd} depicts the singular values of 78 price matrices $\mathbf{Z}\in\mathbb{R}^{1732\times 168}$. The figure shows that singular values decay quickly, and retaining the top 20 could possibly express most of the information in market data. Such an observation not only justifies the trace norm regularization in \eqref{eq:regularization3}, but also hints at fixing $R$ to 20 for a good complexity-performance tradeoff. Figure~\ref{subfig:Psvd} shows the singular values of matrices $\hat{\mathbf{P}}\in\mathbb{R}^{1732\times 168}$ as obtained by solving \eqref{eq:problem}. Interestingly, even though parameter $R$ was set to 20, the rank of $\hat{\mathbf{P}}$'s is no more than 10 in all 78 predictions. 

\color{black}


\begin{figure}
\centering
\includegraphics[width=0.5\textwidth]{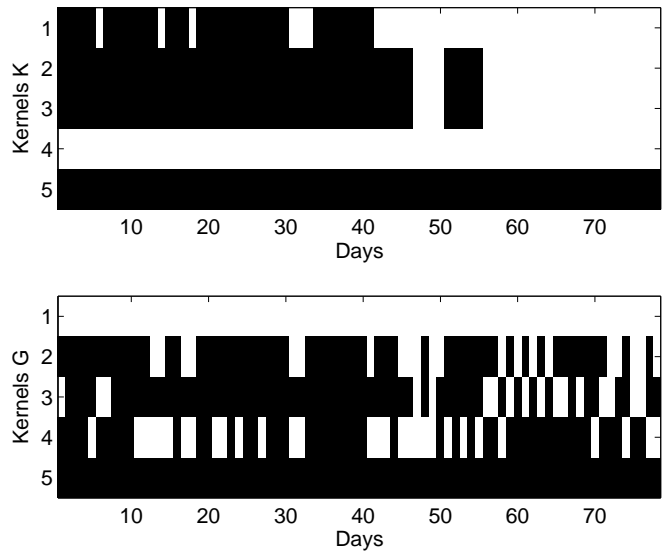}
\caption{Kernel selection: a black (white) square indicates that the respective kernel has been selected (eliminated) while forecasting that specific day.}
\label{fig:kersel}
\end{figure}

\color{black}
Figure~\ref{fig:kersel} shows the kernel selection capability of the novel multi-kernel learning approach. Checking whether the $\{\|\mathbf{B}_l\|_{\mathbf{K}_l}\}_{l=1}^L$ and $\{\|\mathbf{\Gamma}_m\|_{\mathbf{G}_m}\}_{m=1}^M$ obtained by Alg.~\ref{alg:BCD} are zero or not, indicates whether the corresponding kernels, $\{\mathbf{K}_l\}$ and $\{\mathbf{G}_m\}$ have been eliminated. A black (white) square in Fig.~\ref{fig:kersel} indicates that the respective kernel has been selected (eliminated) while forecasting that specific day. Regarding nodal kernels, note that interestingly the identity kernel $\mathbf{K}_4=\mathbf{I}_{1732}$ has been eliminated; hence, providing experimental evidence that coupling price forecasting across CPNs is beneficial. On the other hand, kernel $\mathbf{K}_5$ computed as the sample nodal covariance across the training period seems to capture rich information of CPN pair similarities and is always selected. As far as time kernels are concerned, note that the bandwidth for the Gaussian kernel $\mathbf{G}_1$ turns out to be inappropriate, while the linear kernel $\mathbf{G}_5$ is consistently activated.
\color{black}


\begin{figure}
\centering
\includegraphics[width=0.5\textwidth]{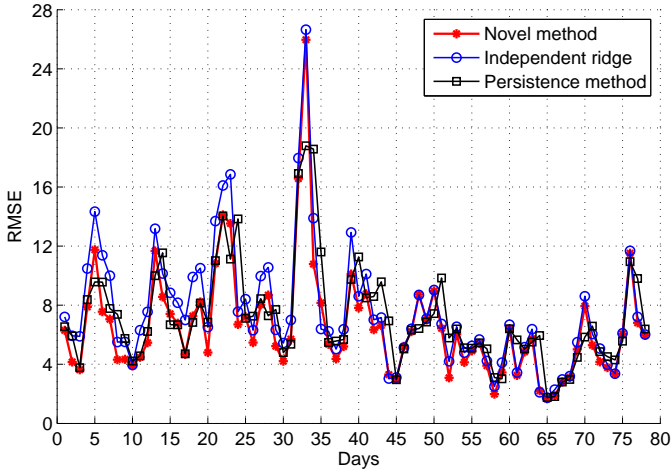}
\caption{RMSE comparison of forecasting methods.}
\label{fig:rmse}
\end{figure}

\color{black}
Finally, the forecasting performance of the novel method is provided in Fig.~\ref{fig:rmse}. Specifically, three methods were tested: (i) the novel multi-kernel learning method; (ii) the ridge regression forecast where each CPN predictor is independently obtained by solving $\min_{\mathbf{a}}~\|\mathbf{z}-\mathbf{G}_1\mathbf{a}\|_2^2 
+ \mu\mathbf{a}^T\mathbf{G_1}\mathbf{a}$ for the Gaussian kernel $\mathbf{G}_1$; and (iii) the persistence method which simply repeats yesterday's prices. The derived low-rank and sparsity-leveraging multi-kernel forecast attains almost consistently the lowest RMSE. The RMSEs averaged across 78 days of the evaluation period are $6.53$, $7.55$, and $7.20$ for the three methods, respectively.
\color{black}

\section{Conclusions}\label{sec:conclusions}
A novel learning approach was developed here for electricity market inference. The congestion mechanisms causing the variations in whole-sale electricity prices were specifically accounted for. After viewing prices across CPNs and hours as entries of a matrix, a pertinent low-rank model was postulated. Its factors were selected from a set of candidate kernels by solving a non-convex optimization problem. Stationary points of this problem can be attained using a computationally attractive block-coordinate descent algorithm. The block-sparse properties of the per-coordinate minimizations facilitate kernel selection. Meaningful nodal kernels were built upon utilizing the related LBA connectivity graph. Applying the novel approach to MISO market data demonstrated its low-rank and kernel selection features. Even though the devised market predictor was based only on publicly available data which may not fully characterize the market outcome, it outperforms standard per-CPN predictors.

\textcolor{black}{The developed kernel selection methodology is sufficiently generic. It can be engaged in any low-rank collaborative filtering setup where kernels need to be selected across two types of features. Extensions to low-rank tensor scenarios where kernels are chosen over three or more feature types is an interesting research direction too. Focusing on applications for smart grids, kernel learning for low-rank models could be used to predict load demand, as well as solar and wind energy, across nodes and time periods.}

\appendix\label{sec:appendix}

\color{black}
\subsection{Proof of Proposition~\ref{pro:subset}}\label{subsec:subset}
\begin{IEEEproof}[Proof of Proposition \ref{pro:subset}] The proof follows the Pareto efficient argument of \cite[App.~A]{Cara10}.
Let $\mathcal{S}_{\lambda}$ and $\mathcal{S}_{\mu}$ be the sets of functions minimizing \eqref{eq:regularization2.5} and \eqref{eq:regularization3} for all $\lambda\geq 0$ and $\mu\geq 0$, respectively. Since $\eqref{eq:regularization2.5}$ is a convex problem, the set $\mathcal{S}_{\lambda}$ coincides with the set of \emph{weakly efficient} functions $\mathcal{S}_p$~\cite{Cara10}: A function $p^*$ belongs to $\mathcal{S}_p$ if at least one of the following conditions hold: 
\begin{enumerate}
\item $p^*\in\arg\min_{p\in\mathcal{P}}~\|\mathbf{Z}-\mathbf{P}\|_F^2$;
\item $p^*\in\arg\min_{p\in\mathcal{P}}~\|p\|_*$;
\item $p^*$ is Pareto efficient, i.e., there is no $p'\in\mathcal{P}$ such that $\|\mathbf{Z}-\mathbf{P}'\|_F^2\leq \|\mathbf{Z}-\mathbf{P}\|_F^2$ and $\|p'\|_*\leq \|p\|_*$ with at least one strict inequality.
\end{enumerate}

Observe next that if $p^*_{\mu}$ minimizes \eqref{eq:regularization3} for some $\mu\geq 0$, then it is also weakly efficient. Hence, $\mathcal{S}_{\mu}\subseteq \mathcal{S}_p=\mathcal{S}_{\lambda}$, which proves the claim.
\end{IEEEproof}
\color{black}

\subsection{Proof of Lemma~\ref{le:sqrt}}\label{subsec:sqrt_proof}
Proving Lemma~\ref{le:sqrt}, requires the following result.

\color{black}
\begin{lemma}\label{le:equal}
If $\left\{f_r^*,g_r^*\right\}_{r=1}^R$ are the minimizers of \eqref{eq:sqrt}, it holds that $\sum_{r=1}^R \|f_r^*\|_{\mathcal{K}}^2 = \sum_{r=1}^R\|g_r^*\|_{\mathcal{G}}^2$.
\end{lemma}

\begin{IEEEproof}[Proof of Lemma \ref{le:equal}]
Arguing by contradiction, suppose there exist $\left\{f_r^0,g_r^0\right\}_{r=1}^R$ minimizing \eqref{eq:sqrt} with $\sum_{r=1}^R \|f_r^0\|_{\mathcal{K}}^2 \neq \sum_{r=1}^R\|g_r^0\|_{\mathcal{G}}^2$. Without loss of generality, assume $\sum_{r=1}^R \|f_r^0\|_{\mathcal{K}}^2 = (1+\epsilon)^2 \cdot \sum_{r=1}^R\|g_r^0\|_{\mathcal{G}}^2$ for some $\epsilon>0$. The minimum value attained in \eqref{eq:sqrt} is $(2+\epsilon)\cdot\sqrt{\sum_{r=1}^R\|g_r^0\|_{\mathcal{G}}^2}/2$.

Consider now the functions $\left\{(1+\epsilon/2)^{-1} \cdot f_r^0\right\}_{r=1}^R$ and $\left\{(1+\epsilon/2) \cdot g_r^0\right\}_{r=1}^R$ which are feasible for \eqref{eq:sqrt}, yielding a cost of $\left(\frac{1+\epsilon}{1+\epsilon/2}+1+\tfrac{\epsilon}{2}\right)\cdot \sqrt{\sum_{r=1}^R\|g_r^0\|_{\mathcal{G}}^2}/2$. The fact that $\frac{1+\epsilon}{1+\epsilon/2}+1+\tfrac{\epsilon}{2}<2+\epsilon$ for all $\epsilon>0$ contradicts the assumed optimality of $\left\{f_r^0,g_r^0\right\}$.
\end{IEEEproof}
\color{black}

\begin{IEEEproof}[Proof of Lemma \ref{le:sqrt}]
Every $p\in \mathcal{P}$ admits a spectral factorization $p(n,t)=\sum_{r=1}^{\infty} \sigma_r u_r(n) v_r(t)$, where $\{\sigma_r\}$ is a non-negative sequence converging to zero, and $\{u_r(n)\} $ and $\{v_r(t)\}$ are orthonormal functions in $\mathcal{N}$ and $\mathcal{T}$, accordingly. The trace norm of $p$ is then defined as $\|p\|_*:=\sum_{r=1}^{\infty} \sigma_r$~\cite{Aber09}.

To show that $h(p)\leq \sqrt{\|p\|_*}$, consider the spectral decomposition of $p=\sum_{r=1}^{R} \sigma_r u_r v_r$. Choose $f_r=\sqrt{\sigma_r}u_r$ and $g_r=\sqrt{\sigma_r}v_r$ for $r=1,\ldots,R$. Since $\{f_r,g_r\}$ are feasible for \eqref{eq:sqrt} and attain a cost of $\sqrt{\|p\|_*}$, it follows that $h(p)\leq \sqrt{\|p\|_*}$.

It is next shown that $\sqrt{\|p\|_*}\leq h(p)$. Because the square root is strictly increasing, it can be applied on \eqref{eq:tracenorm} to yield
\begin{equation}\label{eq:sqrt2}
\|p\|_*^{\frac{1}{2}}{=}\min_{\{f_r,g_r\}} \left\{
\sqrt{\frac{1}{2}\sum_{r=1}^R \|f_r\|_{\mathcal{K}}^2 + \|g_r\|_{\mathcal{G}}^2}: p=\sum_{r=1}^R f_r g_r
\right\}.
\end{equation}
Let $\left\{f_r^*,g_r^*\right\}_{r=1}^R$ be minimizers of \eqref{eq:sqrt}. By Lemma~\ref{le:equal}, they yield a minimum of $h(p)=\sqrt{\sum_{r=1}^R \|g_r^*\|_{\mathcal{G}}^2}$. These minimizers are also feasible for \eqref{eq:sqrt2}, while attaining a cost of $\sqrt{\sum_{r=1}^R \|g_r^*\|_{\mathcal{G}}^2}$. Thus, $\sqrt{\|p\|_*}\leq \sqrt{\sum_{r=1}^R \|g_r^*\|_{\mathcal{G}}^2} = h(p)$ that completes the proof.
\end{IEEEproof}

\subsection{Proof of Theorem~\ref{th:mkl}}\label{subsec:mkl_proof}
Theorem \ref{th:mkl} builds upon the key result of \cite[p.~352-53]{Aronszajn}:
\begin{theorem}[Aronszajn, 1950]\label{th:Aronszajn}
If $K_l$ is the kernel of the function family $\mathcal{H}_{\mathcal{K}_l}$ having norm $\|\cdot\|_{\mathcal{K}_l}$, then $K=\sum_{l=1}^L \theta_l K_l$ for any $L\geq 2$ and $\theta_l>0$, is the reproducing kernel of the function family $f=\sum_{l=1}^L f_l$ with $f_l\in \mathcal{H}_{\mathcal{K}_l}$, having the norm $\|f\|_{\mathcal{K}}^2=\min\left\{\sum_{l=1}^L \frac{\|f_l\|_{\mathcal{K}_l}^2}{\theta_l}:f=\sum_{l=1}^L f_l,f_l\in\mathcal{H}_{\mathcal{K}_l}  \right\}$.
\end{theorem}

\begin{IEEEproof}[Proof of Theorem \ref{th:mkl}]
Theorem \ref{th:Aronszajn} asserts that a conic combination of kernels defines a function family whose members can be alternatively represented as a sum of functions defined by the constituent kernels. Applying this result to the convex combinations of \eqref{eq:convexhulls}, allows replacing \eqref{eq:double} with 
\begin{equation}\label{eq:double2}
\min_{\mathcal{K},\mathcal{G}}\min_{p\in \mathcal{P}'}~ Q(\mathcal{K},\mathcal{G},p)
\end{equation}
where $\mathcal{P}'$ has been defined in \eqref{eq:Pprime}. Upon exchanging the order of minimizations in \eqref{eq:double2}, consider solving the inner one, that is  $\min_{\mathcal{K},\mathcal{G}}~ Q(\mathcal{K},\mathcal{G},p)$. The LS term is constant for a fixed $p\in \mathcal{P}'$, while the two regularization terms can be separately minimized over $\mathcal{K}$ and $\mathcal{G}$, respectively.

Focus now on solving $\min_{\mathcal{K}} \left(\sum_{r=1}^R \|f_r\|_{\mathcal{K}}^2\right)^{\frac{1}{2}}$. By Theorem~\ref{th:Aronszajn}, for a fixed $f_r\in\mathcal{H}_{\mathcal{K}}$, there exist $\{f_{lr}\in \mathcal{H}_{\mathcal{K}_l}\}_{l=1}^L$ such that
\begin{equation}\label{eq:fr_norm}
\|f_r\|_{\mathcal{K}}^2=\sum_{l=1}^L \frac{\|f_{lr}\|_{\mathcal{K}_l}^2}{\theta_l}.
\end{equation}
Summing \eqref{eq:fr_norm} over $r$ and defining $\alpha_l^2:=\sum_{r=1}^R \|f_{lr}\|_{\mathcal{K}_l}^2$ yields
\begin{equation}\label{eq:fr_norm2}
\sum_{r=1}^R \|f_r\|_{\mathcal{K}}^2=\sum_{r=1}^R\sum_{l=1}^L \frac{\|f_{lr}\|_{\mathcal{K}_l}^2}{\theta_l}= \sum_{l=1}^L \frac{\alpha_l^2}{\theta_l}.
\end{equation}
Recall that minimizing over $\mathcal{K}$ amounts to finding the optimum $\{\theta_l\}_{l=1}^L$. By applying the Cauchy-Schwarz inequality, it can be shown that \cite[Lemma~26]{Mich05}
\begin{equation}\label{eq:cauchy}
\min_{\{\theta_l\}_{l=1}^L} \left\{ \sqrt{
\sum_{l=1}^L \frac{\alpha_l^2}{\theta_l}}:
\theta_l>0,
\sum_{l=1}^L \theta_l=1\right\}=\sum_{l=1}^L \alpha_l.
\end{equation}
Utilizing \eqref{eq:cauchy} to minimize the square root of \eqref{eq:fr_norm2}, and replicating the analysis for $\{g_r\}_{r=1}^R$ completes the proof.
\end{IEEEproof}

\subsection{Proof of Lemma \ref{le:canonical}}\label{subsec:canonical_proof}
Lemma \ref{le:canonical} generalizes \cite[Corollary~2]{Puig11} to matrix variables.
\begin{lemma}[\cite{Puig11}]\label{le:MTSO}
The solution to the $\ell_2$-penalized LS problem
\begin{equation*}
\hat{\boldsymbol{\theta}}:=\arg\min_{\boldsymbol{\theta}}~\|\mathbf{y}-\mathbf{X}\boldsymbol{\theta}\|_2^2+\mu \|\boldsymbol{\theta}\|_{2}
\end{equation*}
is $\hat{\boldsymbol{\theta}}=\left(\mathbf{X}^{\top}\mathbf{X} + \frac{\mu^2}{4\hat{w}}\mathbf{I}\right)^{-1}\mathbf{X}^{\top}\mathbf{y}$ when $\|\mathbf{X}^{\top}\mathbf{y}\|_2>\mu/2$; and $\mathbf{0}$, otherwise. The scalar $\hat{w}>0$ minimizes the convex problem
\begin{equation}\label{eq:univariate_hero}
\min_{w\geq 0}~ w - \mathbf{y}^{\top}\mathbf{X} \left(\mathbf{X}^{\top}\mathbf{X} + \frac{\mu^2}{4w}\mathbf{I}\right)^{-1} \mathbf{X}^{\top}\mathbf{y}.
\end{equation}
\end{lemma}

\begin{IEEEproof}[Proof of Lemma \ref{le:canonical}]
Since $\mathbf{B}\succ \mathbf{0}$, the problem in \eqref{eq:canonical} can be equivalently expressed in terms of $\mathbf{X}':=\mathbf{B}^{1/2}\mathbf{X}$ as
\begin{equation}\label{eq:canonical2}
\min_{\mathbf{X}'}\|\mathbf{A}-\mathbf{B}^{1/2}\mathbf{X}'\mathbf{C}^{\top}\|_F^2 + \mu \|\mathbf{X}'\|_F.
\end{equation}  
Upon defining $\mathbf{a}:=\vectorize(\mathbf{A})$ and using property (P), \eqref{eq:canonical2} can be expressed in terms of $\mathbf{x}': = \vectorize(\mathbf{\mathbf{X}'})$ as
\begin{equation}\label{eq:canonical3}
\min_{\mathbf{x}'}~  \|\mathbf{a}-(\mathbf{C}\otimes \mathbf{B}^{1/2})\mathbf{x}'\|_2^2 + \mu \|\mathbf{x}'\|_2. 
\end{equation}
By Lemma~\ref{le:MTSO}, the minimizer of \eqref{eq:canonical3} is the solution of 
\begin{equation}\label{eq:x_vec}
 \left(\mathbf{C}^{\top}\mathbf{C}\otimes \mathbf{B} + \frac{\mu^2}{4w}\mathbf{I}\right)\hat{\mathbf{x}}' =(\mathbf{C}^{\top}\otimes \mathbf{B}^{1/2})\mathbf{a}
\end{equation}
when $\|(\mathbf{C}^{\top}\otimes \mathbf{B}^{1/2})\mathbf{a}\|_2>\mu/2$; or $\hat{\mathbf{x}}'=\mathbf{0}$, otherwise. Using property (P) and if $\hat{\mathbf{x}}'=\vectorize(\hat{\mathbf{X}}')$, then $\hat{\mathbf{X}}'$ satisfies $\mathbf{B}\hat{\mathbf{X}}'\mathbf{C}^{\top}\mathbf{C} + \mu^2/(4w)\hat{\mathbf{X}}'=\mathbf{B}^{1/2}\mathbf{A}\mathbf{C}$ when $\| \mathbf{B}^{1/2}\mathbf{A}\mathbf{C}\|_F>\mu/2$; otherwise, $\hat{\mathbf{X}}'=\mathbf{0}$. Transforming back to the sought $\hat{\mathbf{X}}=\mathbf{B}^{-1/2}\hat{\mathbf{X}}'$, yields finally \eqref{eq:sylvester}.

The scalar $\hat{w}$ in \eqref{eq:sylvester} is the minimizer of the optimization problem obtained after replacing $\mathbf{X}$ and $\mathbf{y}$ in  \eqref{eq:univariate_hero} by $\mathbf{C}\otimes \mathbf{B}^{1/2}$ and $\mathbf{a}$, respectively. Given the singular value decompositions $\mathbf{C}= \mathbf{U}_C \mathbf{\Sigma}_C \mathbf{V}_C^{\top}$ and $\mathbf{B}^{1/2}=\mathbf{U}_B\mathbf{\Sigma}_B\mathbf{V}_B^{\top}$, and after some algebraic manipulations, $\hat{w}$ can be shown to be the minimizer of
\begin{equation}\label{eq:univariate_proof}
\min_{w>0} ~ w - \mathbf{w}^{\top}\left(\mathbf{\Sigma}_C^2 \otimes \mathbf{\Sigma}_B^2 \right) \left(\mathbf{\Sigma}_C^2 \otimes \mathbf{\Sigma}_B^2 + \frac{\mu^2}{4w}\mathbf{I} \right)^{-1}\mathbf{w}
\end{equation}
where $\mathbf{w}:=(\mathbf{U}_C^{\top}\otimes \mathbf{U}_B^{\top})\mathbf{a}$. Recognizing that the matrices in \eqref{eq:univariate_proof} are diagonal and that the $d_1\times d_2$ matrix version of $\mathbf{w}$ is $\mathbf{W}=\mathbf{U}_B^{\top}\mathbf{A}\mathbf{U}_C$, yields \eqref{eq:univariate} thus completing the proof.
\end{IEEEproof}

\bibliographystyle{IEEEtranS}
\bibliography{IEEEabrv,power}
\end{document}